\documentclass[letterpaper]{article}
\usepackage{arxiv} 
\usepackage{times} 
\usepackage{helvet} 
\usepackage{courier} 
\usepackage[hyphens]{url} 
\usepackage{graphicx}
\usepackage{natbib}
\usepackage{caption}
\usepackage{algorithm}
\usepackage{algorithmic}

\usepackage{amsmath}
\usepackage{multirow}
\usepackage{booktabs}
\usepackage{tabularx}
\usepackage{soul}
\usepackage{amssymb}
\nocopyright

\title{ProGraph: Temporally-alignable Probability Guided Graph Topological Modeling for 3D Human Reconstruction}
\author{
    Hongsheng Wang\textsuperscript{\rm 1,2},
    Zehui Feng\textsuperscript{\rm 2},
    Tong Xiao\textsuperscript{\rm 2},
    Genfan Yang\textsuperscript{\rm 2},
    Shengyu Zhang\textsuperscript{\rm 1,\thanks{Corresponding author: sy\_zhang@zju.edu.cn}},\\
    Fei Wu\textsuperscript{\rm 1},
    Feng Lin\textsuperscript{\rm 3}
}
\affiliations{
    \textsuperscript{\rm 1}Zhejiang University, \textsuperscript{\rm 2}Zhejiang Lab  \textsuperscript{3}Fuyao Universiy of Science and Technology

}

\usepackage{bibentry}

\begin{document}

\maketitle

\begin{abstract}
    Current 3D human motion reconstruction methods from monocular videos rely on features within the current reconstruction window, leading to distortion and deformations in the human structure under local occlusions or blurriness in video frames. To estimate realistic 3D human mesh sequences based on incomplete features, we propose Temporally-alignable \textbf{Pro}bability Guided \textbf{Graph} Topological Modeling for 3D Human Reconstruction (ProGraph). For missing parts recovery, we exploit the explicit topological-aware probability distribution across the entire motion sequence. To restore the complete human, \textbf{G}raph \textbf{T}opological \textbf{M}odeling (GTM) learns the underlying topological structure, focusing on the relationships inherent in the individual parts. Next, to generate blurred motion parts, \textbf{T}emporal-alignable \textbf{P}robability \textbf{Dist}ribution (TPDist) utilizes the GTM to predict features based on distribution. This interactive mechanism facilitates motion consistency, allowing the restoration of human parts. Furthermore, \textbf{H}ierarchical \textbf{H}uman \textbf{Loss} (HHLoss) constrains the probability distribution errors of inter-frame features during topological structure variation. Our Method achieves superior results than other SOTA methods in addressing occlusions and blurriness on 3DPW.
\end{abstract}

\section{Introduction}
\label{sec:intro}

3D human pose and shape reconstruction models find applications in various fields such as medical rehabilitation, game development, sports analysis, and clothing design \cite{lupion20243d, du2024joypose, 10443540}. Scenes captured by monocular cameras often exhibit occlusion and depth blur due to motion. One of the significant technical challenges that arise in this context is accurately estimating the 3D human body mesh sequence based on incomplete human features in single-view scenarios. 

3D human body video reconstruction aims to estimate the real 3D human body mesh, and its technical means can be roughly classified into regression-based reconstruction methods and probability distribution-based reconstruction methods. Regression methods focus primarily on frame prediction and lack cross-frame feature dependency modeling without exploiting long-term dependencies.
To reconstruct a realistic 3D human body mesh under occlusion blurred segments, methods based on direct learning of spatial position alignment of 3D human body geometrical structures for direct regression are employed \cite{cho2022FastMETRO, goel2023humans}. These methods can perceive the topological structure of the human body in the current frame. However, these methods make it difficult to predict the motion features of the same sequence over long distances between different frames. When the body moves, occluding, etc., features can be lost, resulting in inaccurate and distorted reconstruction \cite{sengupta2022hierarchical}. The other line of work Hierarchical autoregressively learns the rotation probability distribution of each joint according to the topological structure of the human kinematic tree. This is done by utilizing the distribution to improve the accuracy of joint reconstruction in occlusion-ambiguous scenarios \cite{sengupta2021hierarchical}. However, methods based on probability distribution also have limitations. Since the number of joints in the whole body is relatively sparse, it is difficult to represent the features of the joints in the whole body only through rotation probability distribution, which often leads to distortion and muscle stretching reconstruction results. Therefore, the introduction of topological relationships between vertices of the human body mesh is of vital importance in the probability distribution-based approach.

To align human body topology with 3D spatial features, we introduce a framework named ProGraph. This approach involves creating a probability distribution through human body mesh vertex topology-guided prediction. The framework aims to align features with the 3D geometry of the human body by constructing probability distributions based on feature spaces of the same topology in various frames within a sequence. Specifically, it focuses on capturing potential long-distance motion feature information and addressing missing features to guide the generation of human body topology.

ProGraph is organized into three modules. \textbf{G}raph \textbf{T}opological \textbf{M}odelling (GTM), uses the spatial dimension-based Graph Network module to explicitly learn the topological connections among the vertices of the body mesh and to transform the implicit topological connections within the human body into an explicit representation of a graph structure. This explicit connection ensures that in case of local node loss, it is possible to predict the missing nodes based on the graph structure. The first module can learn the topological associations implicit in the human body from the current frame, but in order to maintain the stability of reconstruction results, the cross-frame feature dependencies implicit in the feature space are also needed, which can be expressed as a probability distribution over the 3D geometry of the human body. Therefore, we developed \textbf{T}emporally-alignable \textbf{P}robability \textbf{Dist}ribution (TPDist), which uses a time-guided latent diffusion module to align the probability distributions of the implied human 3D geometric structures between the adjacent frames in a time-guided manner. This approach is able to predict the probability distribution of the missing features with the aid of cross-frame dependency loss, in order to guide the regression of human topology and construct the coherence of the human mesh sequence. Meanwhile, to hierarchically calculate the error between the predicted results and the true values of different body parts, we proposed \textbf{H}ierarchical \textbf{H}uman \textbf{Loss} (HHLoss), a progressive part-aware spatio-temporal loss to guide perception through the body. Finally, we use a non-parametric model reconstruction method to accurately restore the original human muscle movements.

Our contributions mainly lie in: 

1. To the best knowledge, we make the first attempt to combine the temporally-alignable probability and graph Topology in 3D human pose and shape estimation. The proposed Temporally-alignable Probability Guided Graph Topological Modeling (ProGraph) merges the knowledge from probability distribution and human prior structures, improving our method's accuracy.

2. In ProGraph, we carefully design Hierarchical Human Loss (HHLoss), for accurately calculating the progressive probability error of different body parts.

3. The experimental results reveal the effectiveness of ProGraph in 3D human mesh reconstruction, particularly its ability to excel in challenging scenarios involving occlusions and motion blur. Notably, our approach outperforms the state-of-the-art (SOTA) in video-based reconstruction on the 3DPW \cite{vonMarcard2018} dataset.

\section{Related Work}
There are two main categories of technical means for estimating 3D human pose and shape: regression-based reconstruction methods and probability distribution-based reconstruction methods. Spatial regression based methods \cite{cho2022FastMETRO, lin2023osx} leverage transformer encoders to capture the non-local relationships between human joints and mesh vertices, which facilitates the reconstruction of three-dimensional human mesh from single images. Transformer-based human body reconstruction approaches can generally be categorized into two groups: parameterization methods that employ the SMPL model, and non-parametric methods that directly upsample regression grid vertices. Notably, parameterization approaches, such as SMPLify \cite{bogo2016smpl} and Akash Sengupta \textit{et al}.\cite{sengupta2022hierarchical} have notably demonstrated success in estimating 3D body shape and pose from 2D joints. However, due to their reliance on strong structural priors, these methods may encounter difficulties in cases of severe occlusion or ambiguity, potentially constraining the reconstruction outcomes within a predefined embedding space and hindering alignment with 2D information. In contrast, non-parametric methods offer direct regression of the body shape using forms such as voxels \cite{1240827, 9010852} or the three-dimensional positions of grid vertices \cite{moon2020i2lmeshnet, lin2021mesh}, bypassing the need for parameter prediction. Non-parametric approaches are inherently flexible and often outperform model-based methods in accurately capturing the nuances of human muscle movement. Building on these advancements, our work employs non-parametric methods, achieving results that surpass current state-of-the-art (SOTA) outcomes \cite{zhou2023humaniflow, guo2022extending} in various metrics.

Probability distribution-guided models, particularly those based on Markov chains \cite{song2022denoising, ho2020denoising}, have demonstrated remarkable success in various computer vision applications, such as image generation \cite{nichol2021improved, rombach2022highresolution, dhariwal2021diffusion}, super-resolution \cite{saharia2021image, ho2021cascaded}, and video generation \cite{li2023trackdiffusion, blattmann2023stable, blattmann2023align}. Liu \textit{et al}. \cite{liu2019spatial} introduced Constrained Spatial Smoothing (CSS) for spatial data reconstruction, which has been applied in various domains. Guo \textit{et al}. \cite{guo2022extending} extended regionalization algorithms to address spatial process heterogeneity, providing new insights into spatial data analysis. Diffusion models, capable of learning sample probability distributions, have been applied in innovative ways. For example, Gene Chou \textit{et al}.\cite{li2023diffusionsdf} introduced DiffusionSDF for point cloud completion and reconstruction, predicting real point cloud distributions from sparse samples and achieving impressive results. The Flow method \cite{sengupta2023humaniflow} models human body posture and shape by mapping data to a low-dimensional latent space for representation, generation, and transformation, showing promising predictive capabilities. Similar to our work, DiffMesh \cite{zheng2023diffmesh} generates accurate output mesh sequences by learning the probability distribution of human motion in both the forward and backward processes of the diffusion model. However, it may produce unreasonable mesh outputs in scenes with significant occlusion.

Our contribution, ProGraph, integrates human body topology guidance to enhance hierarchical perception of human joint structures within a frame, mitigating implausible reconstruction outcomes. Additionally, the use of probability distribution models enables our method to effectively track reconstructed characters in dynamic scenes, accurately infer missing structural details between successive frames, and reduce jitter in the reconstruction results.

\section{Method}

\subsection{Overview}

\begin{figure*}[t]
    \centering
   
    \includegraphics[width=1.0\linewidth]{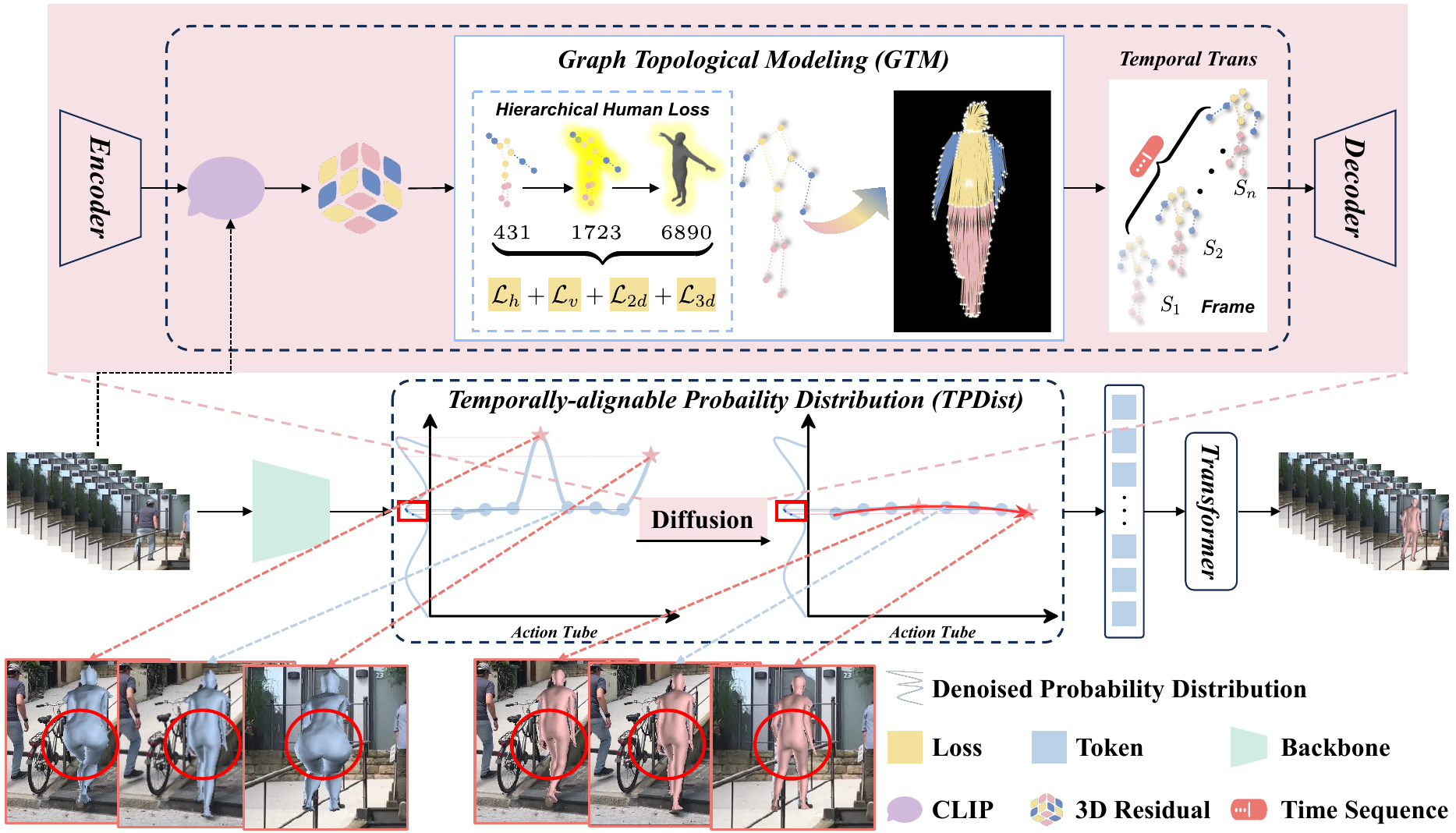}
    \caption{Overview framework. The ProGraph consists of three main components: Temporally-alignable Probability Distribution (TPDist), Graph Topological Modeling (GTM), and Hierarchical Human Loss (HHLoss). Given latent features extracted from the HRNet-W64 backbone, TPDist learns temporally consistent motion features. It leverages the probability distribution of the human body topology within the latent space, as guided by GTM. This enables TPDist to capture information about missing regions due to occlusions. }
    \label{fig:overview}
   
\end{figure*}

Our framework, illustrated in Figure \ref{fig:overview} revolves around three key components: Temporally-aligned Probability Distribution (TPDist), Graph Topological Modeling (GTM), and Hierarchical Human Loss (HHLoss). Given an RGB video as input, GTM progressively encodes the latent space for each frame, capturing the underlying human topological structure. TPDist then leverages GTM to learn the temporal evolution of feature probability distributions for each body part across the entire motion sequence. Subsequently, HHLoss hierarchically constrains the feature errors for each body part, ensuring their consistency. Hierarchically refers to the purpose of the HHLoss: unite errors generated during multi-hierarchical downsampling of GTMs. This synergistic interplay between GTM and TPDist enables accurate 3D body mesh generation, even in scenarios with occluded body parts. The following section details the specific considerations incorporated into our model.

\subsection{Graph Topological Modeling (GTM)}

\begin{figure}[h]
    \centering
    \includegraphics[width=1\linewidth]{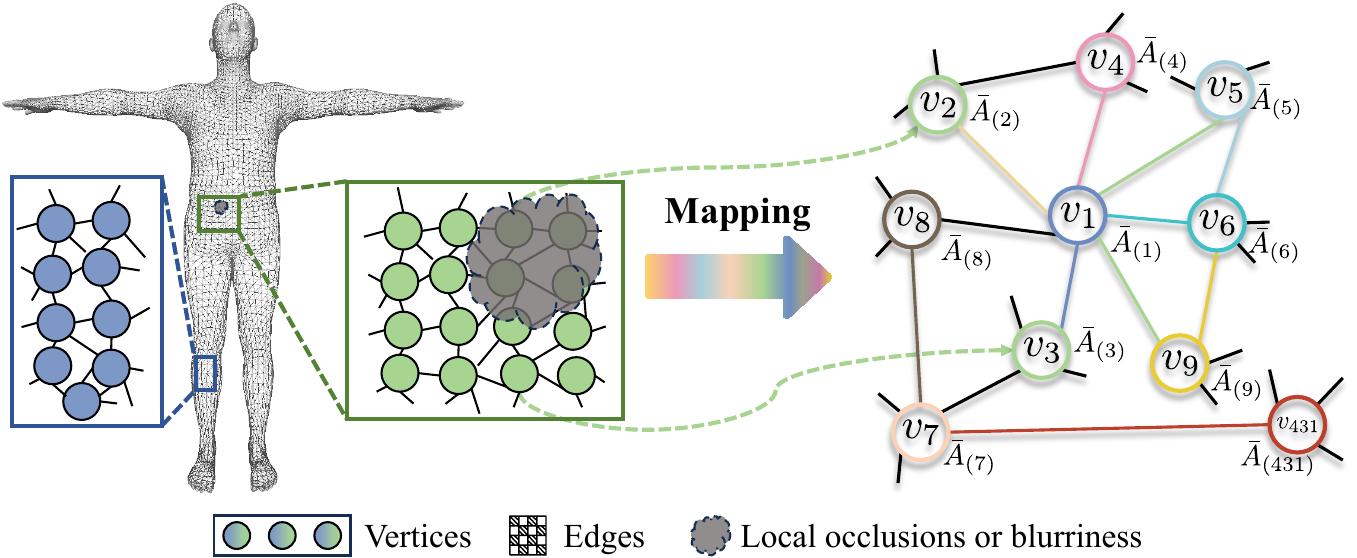}
    \caption{The mapping process between the prior association information of the vertices and the explicit graph structure.}
    \label{fig:mapping}
\end{figure}

Local occlusions within motion sequences frequently result in missing details within the feature space. existing methods often struggle to recover fine details from monocular video inputs, leading to the potential loss of feature details due to localized occlusion \cite{NeRF}. To address this limitation, we introduce Graph Topological Modeling (GTM). This module guides the model to exploit the inherent relationships within the human body's topology.
To effectively leverage the human body's topological structure, GTM performs a dimensionality reduction. It downsamples the 6890 vertices in the SMPL model \cite{2015SMPL} to 431 vertices through linear projection. This focuses on the model's representation of key aspects of the body structure. Subsequently, GTM transforms the implicit connections between these vertices into an explicit graph structure using an adjacency matrix. This process allows GTM to explicitly model the relationships between body parts, rather than relying solely on the individual vertex information. Figure \ref{fig:mapping} illustrates the mapping between the prior association information of the vertices and the resulting explicit graph structure. 

Graph convolution is used to map the explicit graph information to the human body structure in the latent space. For the potential spatial dimension of the latent space \( {( C\times H\times W )} \), each frame of the sequence is a latent space containing the structural features of the target human body \( {Y}={\{Y_1,Y_2,...,Y_n\}} \), \( {Y \in} \)\( \mathbb{R}^{n\times c} \), where \( C \) denotes the latent size, \( H \) and \( W \) represent the height and width of each image frame, and  \( Y \) represents the vector of GraphConv module output, the process of graph convolution is shown in Equation \ref{eq1}:
\begin{align}
\label{eq1}
    {Y}^{\prime}={GraphConv}(\bar{{A}}_\mathbf{Smpl},{Y};{W}_G)=\sigma(\bar{{A}}_\mathbf{Smpl}{Y}{W}_G)
\end{align}
where, \( \bar{{A}}_\mathbf{Smpl} \in\) \( \mathbb{R}^{n\times n} \) is the adjacency matrix of the explicit graph structure, which contains prior information about human body parts and their associations guided by the SMPL model, \( {W}_{G} \) is the trainable parameter, and \( \sigma(\cdot) \) is the activation function that imparts nonlinearity to the network.

\subsection{Temporally-alignable Probability Distribution (TPDist)}
While the (GTM) module enhances the model's resilience against local feature errors, it can still occasionally produce jittering artifacts in the mesh vertices when faced with ambiguous input. To address this limitation, we introduce the Temporally-aligned Probability Distribution (TPDist) module. TPDist tackles the jittering phenomenon and recovers blurred body regions by learning the probability distribution of features for each body part across the entire motion sequence. This comprehensive understanding of feature distribution allows TPDist to predict missing or ambiguous information, leading to smoother and more robust mesh reconstructions. .

TPDist introduces a Temporal Transformer layer after GTM. This layer defines the sequence-aware temporal backbone to align the human topology encoded by GTM frame-by-frame in a temporally consistent manner. Given a sequence data \(x \in \mathbb{R}^{(BT) \times C \times H \times W}\), where \(BT\) is the batch size and time steps, \(C\) is the number of latent channels, \(H\) and \(W\) are the spatial potential dimensions. The TPDist process is shown as follows \cite{GSID:JN5a4RoK1toJ}:

\begin{align}
    x_t = \sqrt{1 - \alpha_{t}} \cdot \varepsilon_{t} + \sqrt{\alpha_{t}} \cdot x_{t-1}
\end{align}

\begin{align}
    x_t = CrossAttention(\gamma_A,x_t)
\end{align}
where \(\varepsilon\) is the Gaussian noise, \(x_t\) is the input to the hidden space at moment \(t\) of the noise addition process, \(\alpha_t\) is a value between [0.0, 1.0] as a coefficient, and \(\gamma_A\) is a latent feature \cite{wang2021actionclip} obtained from the video stream to gradually enhance the action-semantic information of the human body in the process of noise addition.

\begin{gather*}
    {x} \leftarrow {rearrange}({{x_t}},{(B\ T)\ }{C\ H\ W}\rightarrow {{B\ T\ C\ H\ W)}}
\end{gather*}

\begin{align}
    {x}={3DConv}({{x}};{W}_H)=\sigma({{x}}{W}_H)
\end{align}

\begin{align}
x=GraphConv\left(x;\bar{A}_{smpl}\right)
\end{align}
where \(W_H\) is a trainable parameter. It interprets the full noise, \(x_t\) as a batch of independent data with human body topological information. The module encodes it frame-by-frame using GTM to enhance topological information within each frame.

\begin{gather*}
    {x}\leftarrow{rearrange}({{x}},{B\ T\ }{C\ H\ W}\rightarrow{{(B\ H\ W) \ T\ C)}}
\end{gather*}

\begin{align}
    \delta=SelfAttention(x)
\end{align}

To extract temporally consistent features, we synchronize the temporal information with the frame-level topological structure. We achieve this by reorganizing the temporal dimensions to decouple the latent features. This ensures that each resulting structure incorporates information from the entire time series. Subsequently, a Temporal Transformer model is applied along the temporal dimension  \(t\)  to capture the dependencies  \(\delta\)  among identical topological structures across various time steps.

\begin{gather*}
    x_t\leftarrow{rearrange}({x,{{(B\ H\ W)\ T\ C}}\rightarrow{(B\ T)\ }{C\ H\ W})}
\end{gather*}

\begin{align}
    z_t = CrossAttention(\delta, x_t) 
\end{align}

\begin{align}
z_{t-1}=\frac{1}{\sqrt {\alpha_t}}(z_t-\frac{1-\alpha_t}{\sqrt{1-\overline{\alpha_t}}}\varepsilon(z_t,t))+\frac{\sqrt{1-\alpha_t} \sqrt{1-\overline{\alpha}_{t-1}} }{\sqrt{1-\overline{\alpha}_{t} }}I
\end{align}

Following the processing by GTM and the Temporal Transformer layer, the data is reshaped back into video dimensions, where  \(I\)represents a standard Gaussian distribution. Here, the captured dependencies  \(\delta_t\) act as the conditional space, guiding the decoding process within the latent space. This approach enhances the propagation of structural information throughout the sequence. Finally, DDPM \cite{ho2020denoising} process is employed to yield a probability distribution of features that aligns with the 3D geometric structure of the human body. 

\begin{figure}[h]
    \centering
    \includegraphics[width=1.0\linewidth]{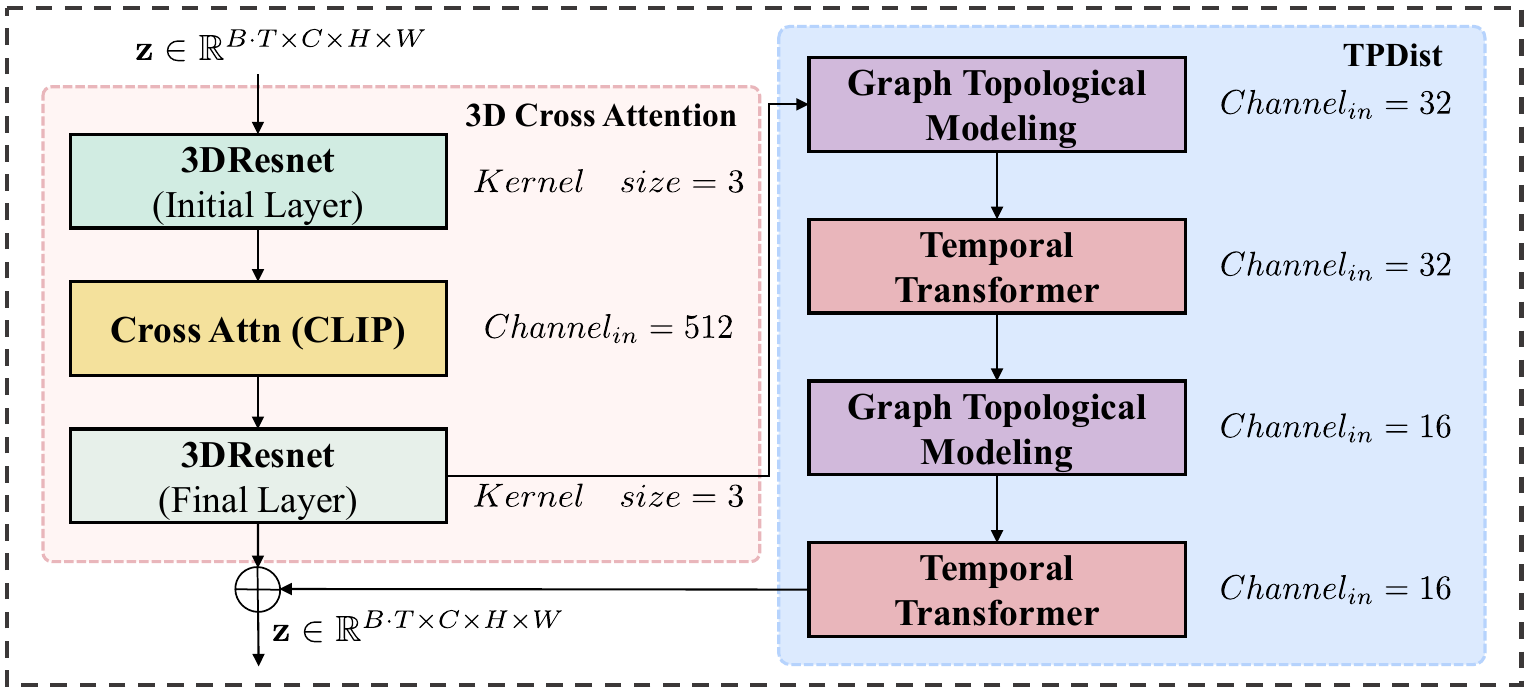}
    \caption{ The ProGraph network architecture integrates a 3D ResNet with CLIP for cross-attention. It also employs a dual-pathway approach, combining Graph Topological Reconstruction and Temporal 3D Transformers. This approach facilitates the alignment of the topological and geometric structures of human body meshes across a video sequence. }
    \label{fig:TPDist}

\end{figure}

To achieve a comprehensive understanding of the human body's structure throughout motion, ProGraph employs the GTM and Temporal Transformer layer iteratively. This two-pass approach constructs a richer graph structure with broader vertex associations, leading to a more complete mapping within the latent space. Consequently, ProGraph is empowered to learn the targeted topological sequences in a spatio-temporally consistent manner. This enhanced learning process translates to improved model accuracy and stability in restoring occluded or blurred body parts. ProGraph achieves this by resampling features from the target probability distribution with the highest confidence scores within the sequences. Figure \ref{fig:TPDist} illustrates the underlying processes of our network. 

\subsection{Hierarchical Human Loss}

In this work, we introduce TPDist, a method that leverages GTM to learn the probability distribution of human body topology. To guarantee consistent topological representation, we propose a novel hierarchical manual guidance approach for the semantic segmentation of 6890 vertices. This approach involves learning the probability distribution of corresponding vertex features at various levels within a region. Subsequently, the error of the vertices is adaptively constrained in a hierarchical manner at each upsampling step. 

Prior to the transformer module, our approach employs FastMetro  \cite{cho2022FastMETRO} to upsample the spatio-temporally aligned human body topology, which incorporates semantic information by subdividing nodes based on body part labels at each step. To avoid spatial redundancy, the sequences are then sampled back to the original resolution (6,890 SMPL mesh vertices) using linear projection. To capture the hierarchical structure and guide the alignment process, we employ softmax pooling to obtain the sequence probability distribution of body parts. Finally, cross-entropy is used to evaluate the difference between the predicted and true distributions for hierarchical topological association alignment.

\begin{align}
\begin{split}
        y=\text{log}\text{softmax}\left(x_{i j }\right)=\log\left(\frac{e^{x_{i j }-c}}{\sum_{k} e^{x_{i j }-c}}\right) \\
    =\left(x_{i j }-c\right)-\log\left(\sum_{k} e^{x_{i j }-c}\right)
\end{split}
\end{align}

\begin{align}
    \mathcal{L}_p\left(y_{\text {pred}}, y_{\text {true}}\right)=y_{\text {true}} \cdot\left(\log y_{\text {true}}-\log y_{\text {pred}}\right)
\end{align}
where \(x\) is the input tensor \(\left(B,N,3\right)\). \(B\) is the size of the batch data, \(N\) is the size of the indexed interval of vertices with well-categorized labels. And \(y\) denotes the probability distribution. \(i\) is the index of the batch, \(j\) is the index of the lattice vertices. We operate along the dimensionality \(N\), with \(c=\max \left(x_{ij }\right)\), \(c\) to constrain the tensor \(x_{ij}\) from exceeding to denote the indexed range of the current lattice vertices.

Human body perception varies across different parts. To account for this, we propose a method that dynamically inherits weights for each body part from the GTM, which reflect the influence of each part on the overall body representation. The weight for a specific part \(\lambda\) is derived from its variance within the GTM, allowing for adaptive tuning of the local influence on the entire body during training. 

Inspired by the detailed vertex classifications of the SMPL model, our method incorporates a fine-grained classification denoted as  \(\mathcal{L}\) and progressively aligns the loss functions associated with the probability distributions for each body part during the upsampling process. This classification scheme utilizes indexed ranges for each label, denoted as  \(\left(s_l,e_l\right)\), where l iterates over the total number of labels  \(m\) .

\begin{align}
    \mathcal{L}_{Human} = \sum_{p=1}^n \lambda_p \mathcal{L}_{p} \cdot \left( \bigvee_{l=1}^{m} (p \geq s_l \land p \leq e_l) \right)
\end{align}

\begin{table*}[t]
\centering
\caption{This table analyzes the influence of the Temporally-Aligned Probability Distribution (TPDist) module and loss functions on 3D human mesh reconstruction performance across the 3DPW and Human3.6M datasets. We evaluate performance using MPVE, MPJPE, and PA-MPJPE metrics. Lower values in these metrics consistently indicate higher accuracy in reconstructing the 3D human mesh. }
\setlength{\tabcolsep}{3pt}
\renewcommand{\arraystretch}{1.0}
\scalebox{0.9}{
\begin{tabular}{cc ccc cc}
\toprule
\multirow{2}{*}{\textbf{TPDist}} & \multirow{2}{*}{\textbf{Loss}} & \multicolumn{3}{c}{\textbf{3DPW}} & \multicolumn{2}{c}{\textbf{Human3.6M}} \\
\cmidrule(lr){3-5} \cmidrule(lr){6-7}
& & \text{MPVE↓} & \text{MPJPE↓} & \text{PA-MPJPE↓} & \text{MPJPE↓} & \text{PA-MPJPE↓} \\
\midrule
×           & ×           & 85.5 & 73.9 & 44.9 & 51.9 & 33.8           \\
\checkmark  & ×           & 82.35 & 72.79 & 44.12 & 49.9 & 33.7           \\
×           & \checkmark  & 81.93 & 72.81 & 44.03 & 49.5 & 33.2           \\
\checkmark  & \checkmark  & \textbf{81.89} & \textbf{72.72} & \textbf{43.82} & \textbf{49.1} & \textbf{33.1}           \\
\bottomrule
\end{tabular}
}

\label{tab:ablation}

\end{table*}

\begin{figure}[t]
    \centering
    \includegraphics[width=1\linewidth]{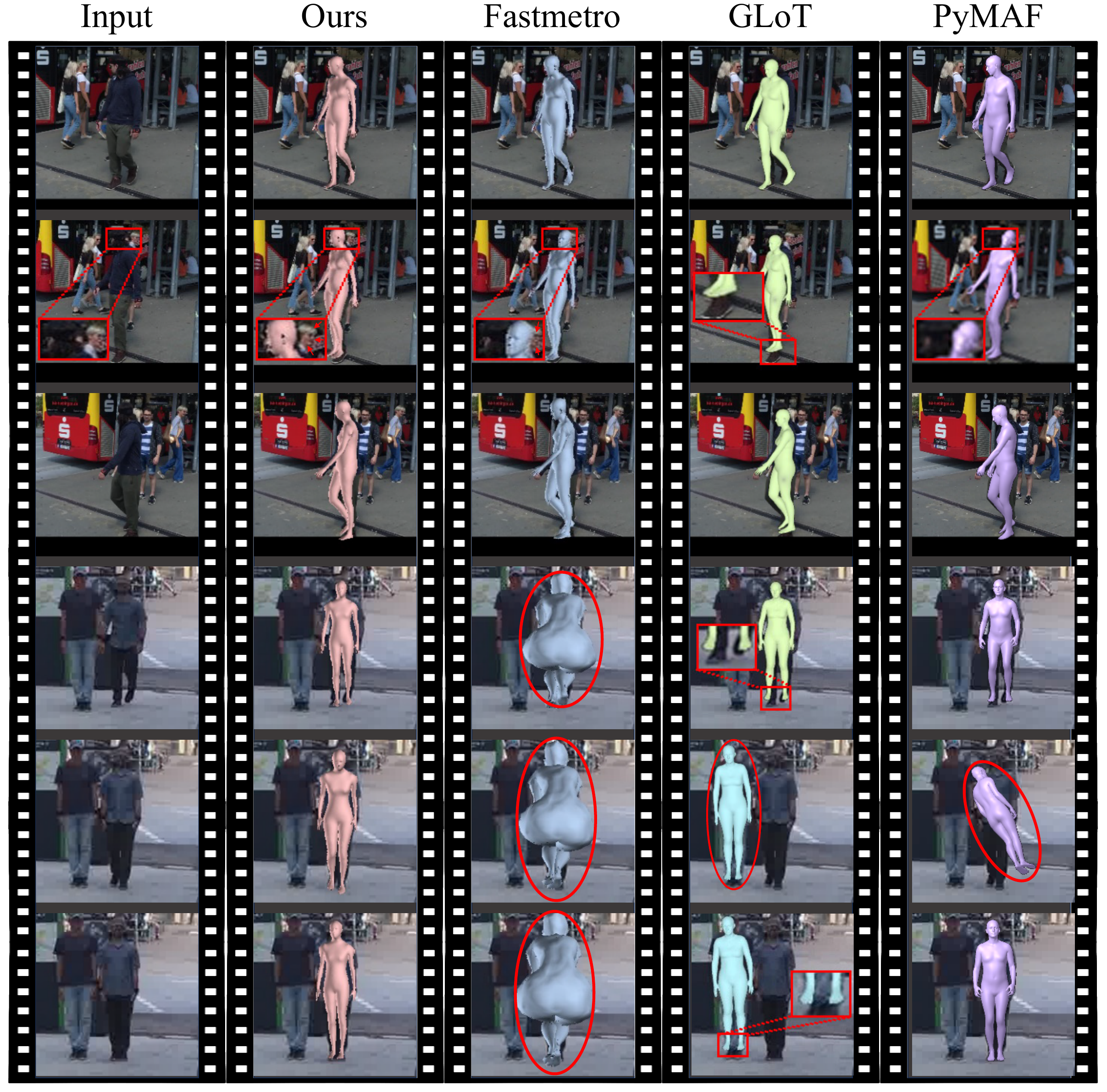}
    \caption{An analysis and comparison of inter-frame prediction results. The original image is presented on the far left, with the outputs of our model, Fastmetro, GLoT, and PyMAF displayed sequentially to the right.}
    \label{fig:interframe}
\end{figure}

\section{Experiments}

\begin{table*}[t]
\centering
\caption{Comparison of frame-based methods and video-based methods on 3DPW and Human3.6M datasets.}
\renewcommand{\arraystretch}{1.0}
\scalebox{0.7}{
\begin{tabular}{lcc ccc cc}
\toprule
\multirow{2}{*}{\textbf{Method}} & \multirow{2}{*}{\textbf{Output-type}} & \multicolumn{3}{c}{\textbf{3DPW}} & \multicolumn{2}{c}{\textbf{Human3.6M}} \\
\cmidrule(lr){3-5} \cmidrule(lr){6-7}
& & \text{MPVPE↓} & \text{MPJPE↓} & \text{PA-MPJPE↓} & \text{MPJPE↓} & \text{PA-MPJPE↓} \\
\midrule
\multicolumn{7}{c}{Frame based} \\
\cmidrule(lr){1-7}
Graphormer\cite{lin2021mesh} & vertices & 87.7 & 74.7 & 45.6 & 51.2 & 34.5 \\
METRO \cite{9578630} & vertices & 88.2 & 77.1 & 47.9 & 54.0 & 36.7 \\
Hybrik-X \cite{li2023hybrik} & parameters & 94.5 & 80.0 & 48.8 & - & - \\
Pymaf-X \cite{zhang2023pymaf} & parameters & 110.1 & 92.8 & 58.9 & 57.7 & 40.5 \\
Potter \cite{zheng2023potter} & parameters & 87.4 & 75.0 & 44.8 & 56.5 & 35.1 \\
Fastmetro \cite{cho2022cross} & vertices & 84.1 & 73.5 & 44.6 & 52.2 & 33.7 \\
PointHMR \cite{kim2023sampling} & vertices & 85.5 & 73.9 & 44.9 & 48.3 & 32.9 \\
\cmidrule(lr){1-7}
\multicolumn{7}{c}{Video based} \\
\cmidrule(lr){1-7}
HMMR \cite{kanazawa2019learning} & parameters & 139.3 & 116.5 & 72.6 & - & 56.9 \\
VIBE \cite{kocabas2020vibe} & parameters & 99.1 & 82.9 & 51.9 & 65.6 & 41.4 \\
TCMR \cite{choi2021beyond} & parameters & 111.3 & 95.0 & 55.8 & 62.3 & 41.1 \\
MAED \cite{wan2021encoder} & parameters & 92.6 & 79.1 & 45.7 & 56.4 & 38.7 \\
MPS-Net \cite{wei2022capturing} & parameters & 109.6 & 91.6 & 54.0 & 69.4 & 47.4 \\
GLoT \cite{shen2023global} & parameters & 96.3 & 80.7 & 50.6 & 67.0 & 46.3 \\
4DHumans \cite{goel2023humans} & parameters & 82.2 & \textbf{70.0} & 44.5 & \textbf{44.8} & 33.6 \\
Ours & vertices & \textbf{82.12} & 72.72 & \textbf{43.82} & 49.1 & \textbf{33.1} \\
\bottomrule
\end{tabular}
}

\label{tab:comparison}

\end{table*}

\begin{figure}[t]
    \centering
    \includegraphics[width=1\linewidth]{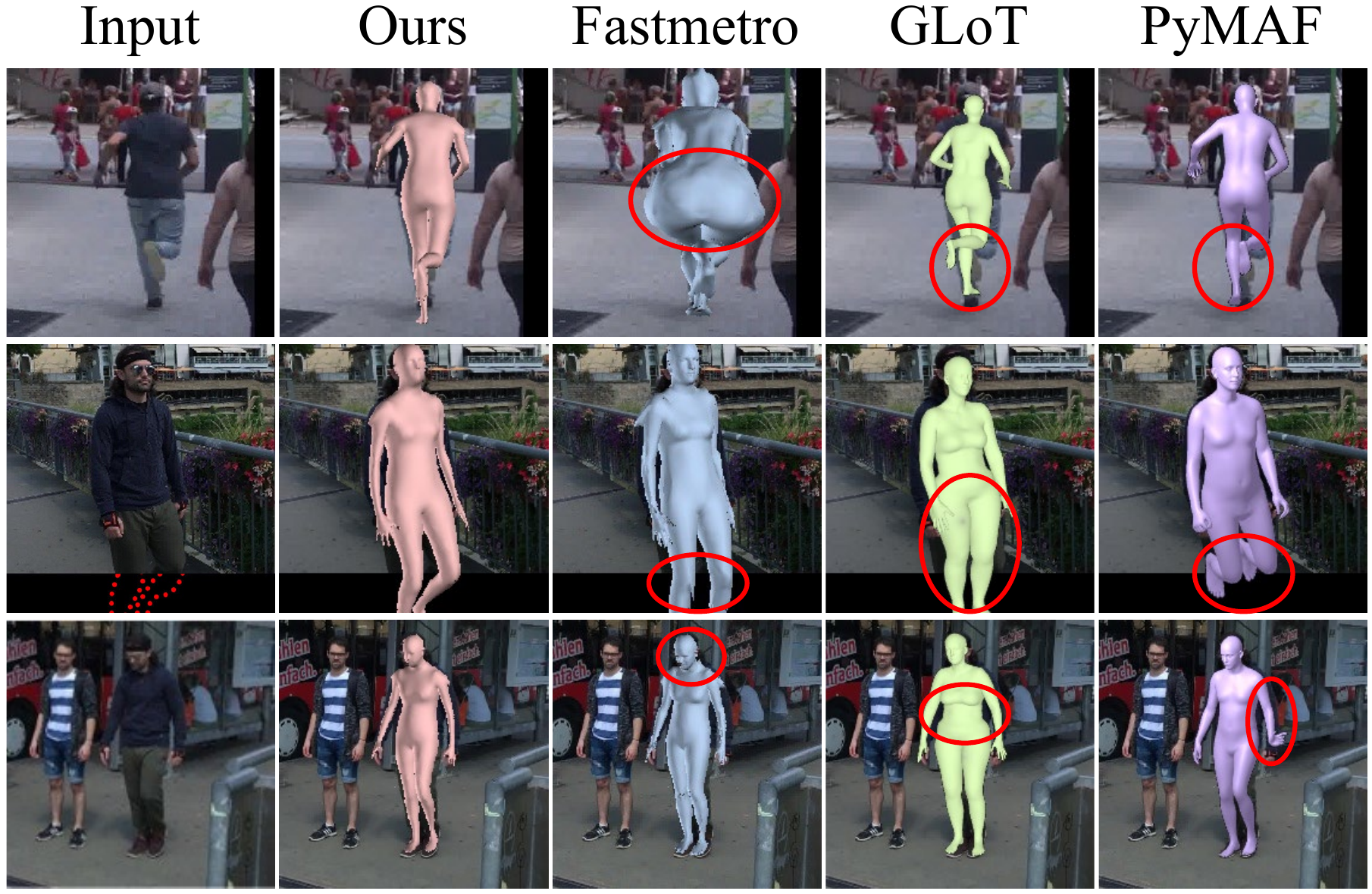}
    \caption{This figure conducts analysis and comparison of intra-frame prediction results, showcasing the capability to handle complex scenes, such as walking people. The original image is displayed on the far left, followed to the right by the outputs of our model, Fastmetro, GLoT, and PyMAF, respectively.}
    \label{fig:intra-frame}
\end{figure}

\begin{figure}[t]
    \centering
    \includegraphics[width=1.0\linewidth]{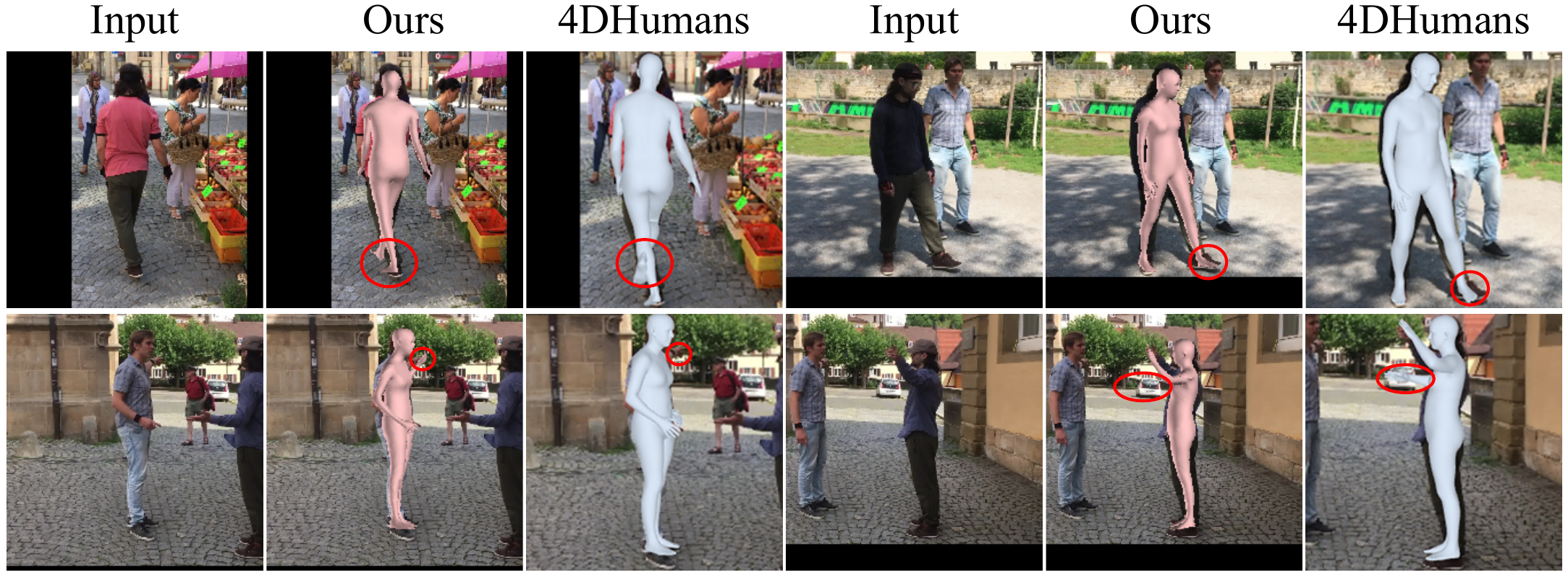}
    \caption{The comparison of our framework and 4DHumans. The red circles around the feet areas emphasize the superior precision of our model's pose estimation, particularly in challenging scenarios with intricate backgrounds or complex poses.}
    \label{fig:4DHumans}
\end{figure}

\subsection{Datasets and metrics}
We leverage publicly available datasets for training our model, including Human3.6M   \cite{HAN201785} and 3DPW \cite{vonMarcard2018}. The 3DPW dataset is used for both training and fine-tuning the model. We evaluate the model's performance on the 3DPW test set. For Human3.6M evaluations, we adhere to the P2 protocol setting. Consistent with prior works  \cite{kanazawa2019learning, kocabas2020vibe, wei2022capturing}, we employ three intra-frame metrics for evaluation: MPJPE, MPJPE, and PA-MPJPE.

\subsection{Ablation Study}
Temporally-alignable Probability Distribution (TPDist) is implemented on the basis of Graph Topological Modeling (GTM), in this section, we will perform two sets of ablation studies, on the 3DPW and Human3.6M datasets to validate the effectiveness of the key modules proposed in our approach. All ablation methods are trained and tested on both datasets, which are widely used benchmarks for 3D human pose and shape reconstruction.

\subsubsection{Efficacy of TPDist.}

TPDist learns the probability distribution of each part, guiding the occlusion and blurring of certain parts for more accurate perception. This allows for efficient extraction of spatial features, thereby improving the quality of 3D human generation. To verify the effectiveness of the module, a set of ablation experiments was conducted. Table \ref{tab:ablation} compares the performance of our model with and without TPDist during training.  The results show that the TPDist reduces the MPJPE and PA-MPJPE metrics by 2.4mm and 0.6mm respectively on the Human3.6M dataset, and by 3.57mm for MPVPE, MPJPE, and PA-MPJPE metrics on the 3DPW dataset. These reductions indicate that TPDist effectively guides the probability distribution between the predicted results and ground truth, leading to more accurate human mesh reconstruction.

\subsubsection{Benefit from HHLoss.}

We compare the method using HHLoss with the baseline. As shown in Table \ref{tab:ablation}, the HHLoss method achieves significant improvements on both the Human3.6M and 3DPW datasets. On Human3.6M, it reduces MPJPE by 1.93mm and PA-MPJPE by 0.1mm. Even more notably, on 3DPW, it reduces MPVPE, MPJPE, and PA-MPJPE by 3.15mm, 1.11mm, and 0.78mm, respectively. These results demonstrate the effectiveness of the HHLoss module's ability to adaptively learn probability distributions at the time level, leading to more accurate 3D human pose estimation. 

\subsubsection{Validating the interaction between TPDist and HHLoss.}

Table \ref{tab:ablation} showcases the significant improvements achieved by combining TPDist and HHLoss compared to a baseline model. On the Human3.6M dataset, our method reduces MPJPE by 2.8mm and PA-MPJPE by 0.7mm. Even more impressive are the results on 3DPW, where MPVPE, MPJPE, and PA-MPJPE are reduced by 3.15mm, 1.11mm, and 0.78mm, respectively. These reductions \textbf{surpass the improvements achievable with each module individually, }demonstrating the synergistic effect of TPDist and HHLoss in enhancing 3D human pose estimation accuracy.

\subsection{Comparative Study}
\subsubsection{Qualitative Analysis.}\textbf{}Compared with the existing methods, ProGraph effectively tackles the challenge  of human body distortion caused by missing or unclear parts of the human body. The following shows the comparison effect: 

Figure \ref{fig:interframe} illustrates the comparative effects of reconstructing different human motion sequences in two scenarios with missing or unclear body parts. The first three rows represent the first motion sequence, while the last three rows represent the second sequence. In the first motion sequence (top rows), both FastPose and PyMAF suffer from severe head distortions. This is likely because the input image has a blurry head region. These methods mistakenly use features from nearby, non-target people to reconstruct the head, leading to inaccuracies.  Similarly to the second motion sequence, GLoT fails to accurately reconstruct the foot position due to ambiguity in the target foot region of the input image. Our ProGraph framework addresses these challenges by learning the probability distribution of each body part across the entire motion sequence. This learning leverages the explicit body topology information extracted from each 2D frame. Even if a particular body part is unclear or blurred in a single frame, ProGraph can still reconstruct it accurately by sampling from the real probability distributions learned from other frames in the sequence.

To evaluate our method's ability to handle incomplete or unclear body parts, we conducted comparative experiments across three scenarios. Figure \ref{fig:intra-frame} showcases the reconstruction results, comparing our approach with existing methods.  The results demonstrate that FastMetro, GLoT, and PyMAF struggle to recover accurate human body mesh information for occluded or missing body parts.  In contrast, our proposed GTM module explicitly captures the body's topology through a graph structure. This allows GTM to learn the intrinsic correlations between different body mesh vertices. Consequently, even in scenarios with occlusions, blurry images, or missing local features, GTM can leverage these correlations to predict the missing information and achieve more accurate reconstructions.

To visually demonstrate ProGraph's effectiveness, we compared it with 4DHumans on the 3DPW dataset  Figure \ref{fig:4DHumans}. In the first line, the reconstruction result of 4DHumans show incorrect estimation of the legs and feet.  Similarly, in the second line of the scene, the torso and upper arm are obscured, leading to incorrect inferences by 4DHumans regarding the arm position.

\subsubsection{Quantitative Analysis.}\textbf{ }

(i) \textbf{ProGraph Outperforms Video-based Methods}. We compare ProGraph with video-based methods on the 3DPW dataset. As shown in Table \ref{tab:comparison}, ProGraph achieves superior performance on most evaluation metrics. Notably, compared to 4DHumans, our method reduces MPVPE and PA-MPJPE by 3.1mm and 0.6mm, respectively. These results establish ProGraph as a new state-of-the-art method for accurate human mesh reconstruction. . 

(ii) \textbf{ProGraph Outperforms Frame-based Methods}. To validate the effectiveness of our framework in 3D human pose and shape reconstruction, we compare ProGraph with previous frame-based methods on the 3DPW dataset using the MPVPE, MPJPE, and PA-MPJPE metrics. As shown in Table \ref{tab:comparison}, we outperform the existing methods on all metrics. For example, ProGraph reduces MPVPE by 3.38 mm, MPJPE by 1.18 mm, and PA-MPJPE by 1.08 mm compared with PointHMR on 3DPW. The effectiveness of our method is demonstrated by the significant advantages of these metrics. Experimental results show that we still have the superiority even on frame-based methods.

\subsection{Visualization Results}

\subsubsection{Visualization of TPDist.}
\begin{figure}[h]
    \centering
    \includegraphics[width=1.0\linewidth]{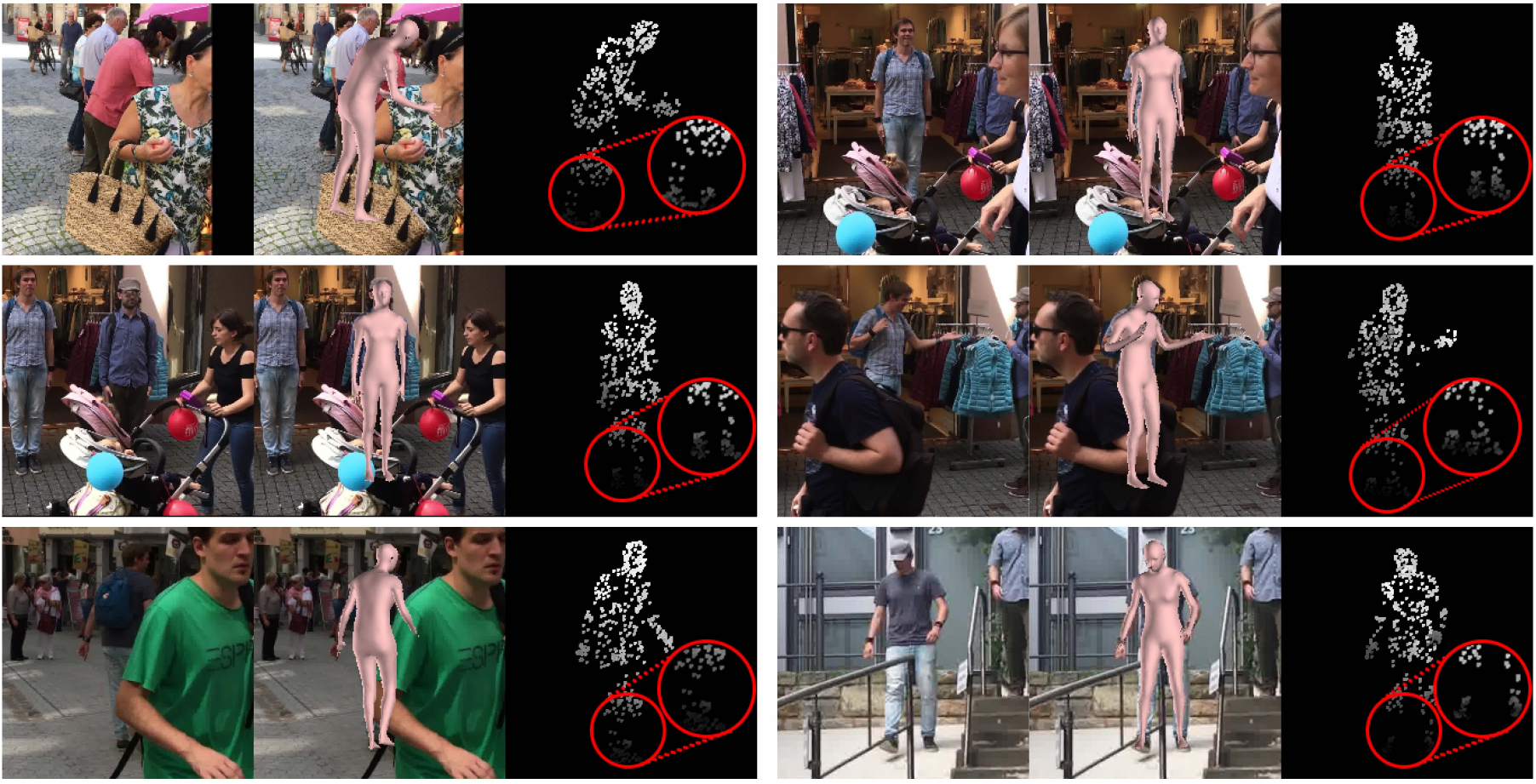}
    \caption{The reconstruction results and probability distribution of human body parts during the prediction process.}
    \label{fig:block}
\end{figure}

As Figure \ref{fig:block} shows, it visualizes the probability distribution of human body parts during the prediction process. The experimental results demonstrate that our method can leverage TPDist to achieve accurate reconstructions, even in complex scenes with occlusions or blurred body parts. 

\subsubsection{Results of wild images.}
\begin{figure}[h]
    \centering
    \includegraphics[width=1.0\linewidth]{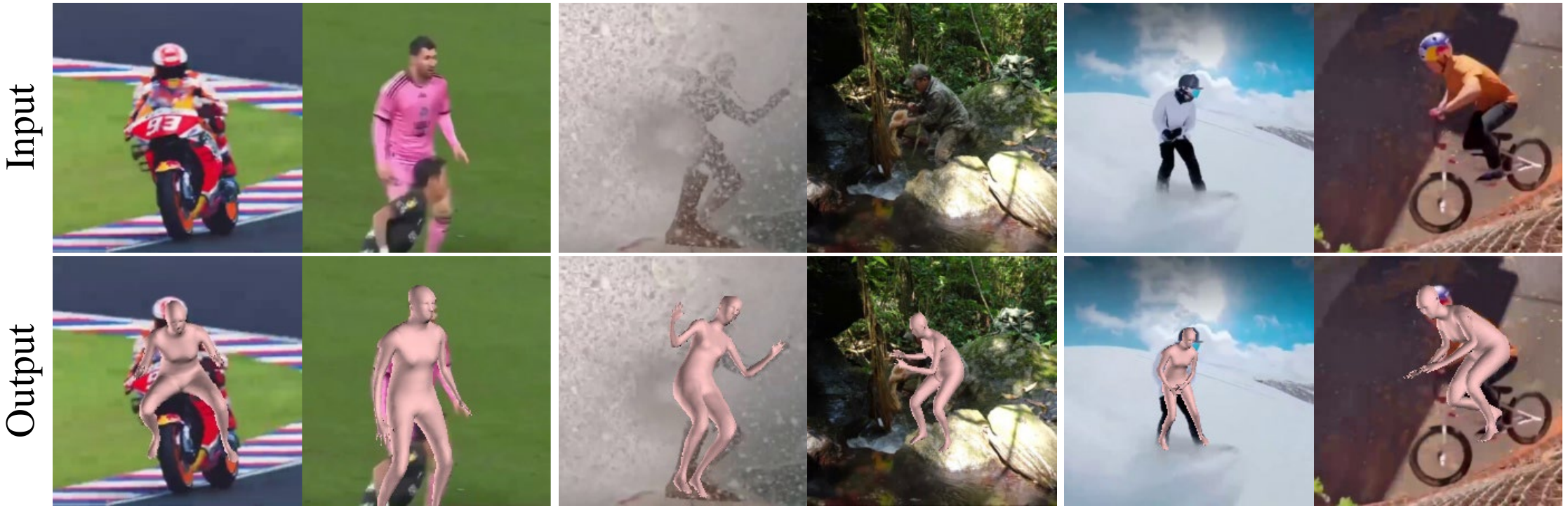}
    \caption{Human body reconstructions by our method. It enables accurate 3D human body reconstruction from videos, even with incomplete parts and occlusions, demonstrating exceptional robustness and high fidelity in 3D pose and shape reconstruction.}
    \label{fig:first}
\end{figure}

Figure \ref{fig:first} shows ProGraph's prediction results for obscured wild images. For wild pictures of complex scenes, ProGraph can effectively  predict the probability distribution of human body parts to generate highly realistic human pose and shape estimations. 

\section{Conclusion}

This work introduces ProGraph, a novel framework for reconstructing 3D human body mesh from monocular videos. ProGraph addresses the challenges of occlusion and blurriness by constructing probability distributions for human body topology across video frames. This significantly improves 3D human pose regression accuracy. Our framework outperforms state-of-the-art (SOTA) methods on 3DPW and Human3.6M, achieving superior reconstructions under challenging conditions. ProGraph's robustness paves the way for various applications requiring accurate 3D pose estimation in unconstrained environments. 

\bibliography{main}

\end{document}